\documentclass[letterpaper]{article}
\usepackage[preprint]{aaai2027}
\usepackage[hyphens]{url}
\usepackage{graphicx}
\usepackage{natbib}
\usepackage{caption}
\usepackage{booktabs}
\usepackage{array}
\usepackage{amsmath}
\usepackage{amssymb}
\ifdefined\blueversion
  \newcommand{\changed}[1]{{\color{blue}#1}}
  \newcommand{\changedcolor}{\color{blue}}
  \newcommand{\changedfigure}[1]{%
    \begingroup
    \setlength{\fboxsep}{0pt}%
    \setlength{\fboxrule}{0.8pt}%
    \fcolorbox{blue}{white}{#1}%
    \endgroup}
\else
  \newcommand{\changed}[1]{#1}
  \newcommand{\changedcolor}{}
  \newcommand{\changedfigure}[1]{#1}
\fi
\newcolumntype{L}[1]{>{\raggedright\arraybackslash}p{#1}}
\newcolumntype{C}[1]{>{\centering\arraybackslash}p{#1}}
\newcolumntype{R}[1]{>{\raggedleft\arraybackslash}p{#1}}
\AtBeginDocument{%
  \setlength{\abovedisplayskip}{5pt plus 1pt minus 1pt}%
  \setlength{\belowdisplayskip}{5pt plus 1pt minus 1pt}%
  \setlength{\abovedisplayshortskip}{3pt plus 1pt}%
  \setlength{\belowdisplayshortskip}{3pt plus 1pt minus 1pt}%
}

\title{CG-World: A Large-Scale World-State Dataset for World Model Research}
\author{
Yiming Cai\textsuperscript{\rm 1,*},
Fangjie Yu\textsuperscript{\rm 1,*},
Meiqing Yu\textsuperscript{\rm 2},\\
Ziyue Shi\textsuperscript{\rm 3},
Yong Guo\textsuperscript{\rm 1},
Pengfei Yuan\textsuperscript{\rm 1}
}
\affiliations{
\textsuperscript{\rm 1}CGyear World Model Lab \quad
\textsuperscript{\rm 2}Tsinghua University \quad
\textsuperscript{\rm 3}University of New South Wales\\
\textsuperscript{*}Corresponding authors:
caiyiming@cgyear.com.cn, yufangjie@cgyear.com.cn
}

\makeatletter
\g@addto@macro\@maketitle{%
\par\centering
\changedfigure{\includegraphics[width=0.96\textwidth]{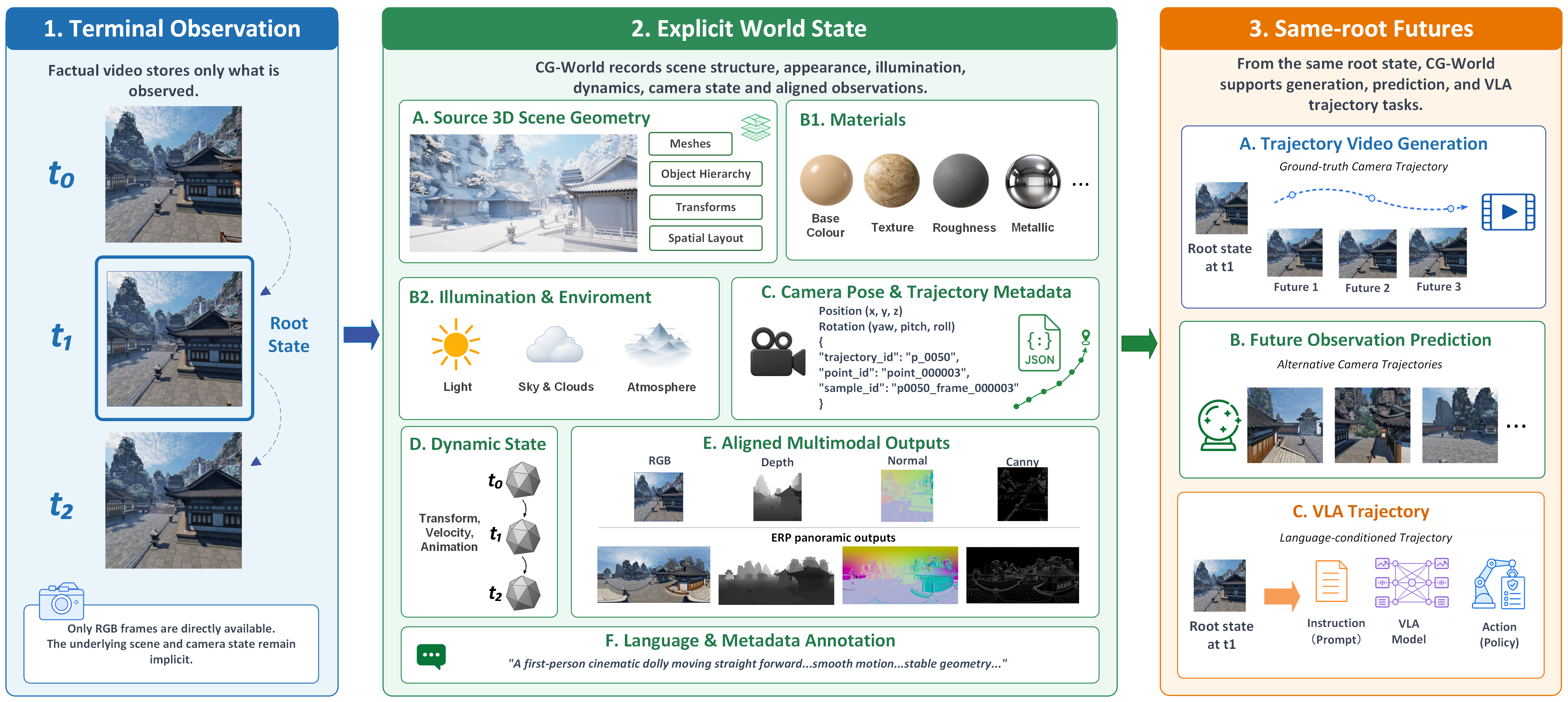}}\par
\captionof{figure}{\changed{CG-World. Unlike conventional datasets that retain only terminal observations, CG-World aligns explicit world states, multimodal observations, and same-root intervention branches, enabling models to learn how the world evolves rather than merely how it appears.}}
\label{fig:overview}\par
}
\makeatother

\begin{document}
\raggedbottom
\maketitle

\begin{abstract}
World models require learning the joint distribution of states, actions, events, and observations, whereas existing public video datasets, robotic trajectories, and simulation data typically cover only a subset of these elements. To address this limitation, we introduce CG-World \changed{(Figure~\ref{fig:overview})}, a world-state data protocol derived from industrial computer graphics production pipelines, together with a corresponding large-scale dataset. The dataset explicitly records intermediate world states, including multimodal semantic information, spatial representations, skeletal and controller states, motion curves, camera and lighting parameters, physics caches, contact events, and multi-pass rendering outputs. CG-World v1 contains approximately 850,000 world-state segments, each ranging from 1 to 5 seconds. It separates underlying states, observations, events, relations, and branch metadata, and organizes these heterogeneous records into temporally aligned data units, thereby providing an explicit and comprehensive representation of world-state information. To further support intervention-based learning and counterfactual reasoning, CG-World introduces a unified branch lineage that categorizes world-state trajectories into factual trajectories, observational interventions, action interventions, mechanism interventions, and strict counterfactual branches. For each branch, the intervention target, invariant variables, and alternative outcomes are explicitly recorded. We evaluate CG-World on three controlled tasks \changed{relevant to world-model research}: geometry-conditioned video generation, action prediction, and closed-loop transfer of vision-language-action policies. Experimental results demonstrate that CG-World provides reusable structured supervision for controlled generation, action modeling, and embodied policy transfer, leading to consistent improvements in model performance. In future work, we aim to further expand the dataset through continued data collection and community collaboration, with the goal of establishing CG-World as a shared data infrastructure for general-purpose world models, Physical AI, and embodied intelligence research.
\end{abstract}


\setcounter{table}{1}

\section{Introduction}

World model research is shifting from video generation toward world understanding and prediction. The field has progressed from compressing observation histories into latent representations to maintaining conditionable, intervenable, and interactive representations in latent space, with the broader goal of jointly modeling state, dynamics, control, and physical regularities \cite{ding2025world}. Kinetics-700, Ego4D, and Panda-70M have supported video learning \cite{carreira2019kinetics,grauman2022ego4d,chen2024panda70m}, but scale alone does not guarantee physical understanding \cite{motamed2026physical}. Public observational data are dominated by video datasets that rarely provide actions, contacts, and related signals aligned with video, limiting their transfer to decision-making models for embodied intelligence and autonomous driving. Robotic datasets such as Open X-Embodiment, BridgeData V2, and DROID \cite{openx2024,walke2023bridgedata,khazatsky2024droid} organize domain-specific observations, actions, and demonstrations, but generally do not expose solver states, contact graphs, material parameters, or comparable intermediate variables. Simulation data offer readable states, controllable parameters, and repeatable processes, yet remain the output of particular engines and configurations and are therefore subject to the reality gap, solver bias, and limited cross-domain generalization \cite{hafner2025dreamerv3,li2020causal}.

More consequential than debates over world model architectures is the absence of a data foundation that supports these diverse research directions. Language and multimodal learning benefit from abundant high-quality corpora \cite{raffel2020t5}. By contrast, public datasets for world models, including video-generation-oriented models, usually cover only a subset of visual sequences, camera poses, depth, panoramic observations, and 3D information. A high-quality public dataset that jointly aligns 3D scene state, panoramic or multiview observations, depth and geometric ground truth, camera trajectories, and final video within the same spatiotemporal sample remains lacking; multicondition datasets that additionally align actions and physical information are even scarcer.

Industrial computer graphics production yields not only final videos but also explicit intermediate records of physics, time, and space within CG engines, providing detailed descriptions of world state. Although these records are not equivalent to real-world physical ground truth, they numerically describe the motion, interactions, and generative processes underlying the observed output and remain aligned with the final video at both entity and temporal levels. They are therefore well suited to multimodal process supervision for world models, yet have not been systematically curated into world model training data.

To address this gap, we follow the precedent of ImageNet \cite{deng2009imagenet}, which organizes images as a hierarchical semantic database. Centered on spatiotemporal alignment, we establish a hierarchical data-governance protocol for describing world states and construct a large-scale dataset and counterfactual branch subset for cross-task world model training and evaluation.

\section{Related Work}

\noindent\textbf{Real-world observational data.} ScanNet \cite{dai2017scannet} and Matterport3D \cite{chang2017matterport3d} provide geometry and camera poses from real RGB-D and 3D scans, offering important support for controllable video generation. However, these data remain centered on terminal observations and generally lack intermediate physical states strictly aligned with temporal observations. Genie, GAIA-1/2, and Cosmos have demonstrated the foundational capabilities of video world models \cite{bruce2024genie,hu2023gaia1,russell2025gaia2,nvidia2025cosmos}. Subsequent work such as PhyWorld improves the physical plausibility of video continuations through continued post-training and physics-preference optimization, further indicating that terminal video observations alone do not provide directly learnable mechanism supervision \cite{zhao2026phyworld}.

\noindent\textbf{Sensor-collected datasets and embodied intelligence.} Open X-Embodiment aggregates more than one million real robot trajectories in a standardized format. Models such as RT-2 \cite{zitkovich2023rt2} and OpenVLA \cite{kim2025openvla} show that jointly training on Internet-scale vision-language knowledge and robot behavior data can improve semantic generalization and cross-task control. Although these datasets capture real sensors, embodiments, and operational noise, they generally cannot record the complete latent state of the world. CG-World is complementary: it supplies dense intermediate-state supervision that is difficult to obtain through real-world data collection.

\noindent\textbf{Synthetic worlds and simulation platforms.} Synthetic data generators and simulators such as Kubric, iGibson, and ThreeDWorld \cite{greff2022kubric,li2022igibson,gan2021threedworld} provide controllable environments, rich annotations, physical simulation, and, in some cases, intervention variables. Yet state protocols and criteria for physical equivalence remain inconsistent across simulators, and different engines or parameter settings may approximate the same physical process with different dynamics. CG-World instead derives from delivered production records and includes substantial expert-authored data unavailable from engine computation alone, such as fluid motion created with 3D particle systems.

\noindent\textbf{Physics and causality.} IntPhys and PHYRE evaluate intuitive physics and interactive reasoning \cite{riochet2022intphys,bakhtin2019phyre}; CLEVRER \cite{yi2020clevrer} unifies description, explanation, prediction, and counterfactual reasoning in a video framework; and CausalWorld \cite{ahmed2021causalworld} enables explicit interventions on causal variables associated with objects and robots in manipulation environments. Most visual-reasoning benchmarks, however, remain limited to discriminative question answering, plausibility judgments, or simplified physical systems and consequently cover a narrow set of application domains.

In summary, world model training still lacks a standardized data system that systematically aligns semantics, actions, physics, observations, and counterfactuals. CG-World addresses this gap by placing explicit intermediate physical states from industrial production, multi-pass visual observations, and same-root branch lineages within a shared spatiotemporal data system.

\section{World-State Data Protocol}

\changed{CG-World defines a logical protocol that separates world-state elements within aligned samples and maps them to industrial or sensor formats. This yields a unified, constructible, transformable, and verifiable representation. Table~\ref{tab:world-state-categories} summarizes its seven categories and representative contents.}

\begin{table*}[!t]
\centering
\footnotesize
\setlength{\tabcolsep}{2.5pt}
\renewcommand{\arraystretch}{1.00}
{\changedcolor
\begin{tabular}{@{}L{0.15\textwidth}L{0.43\textwidth}L{0.36\textwidth}@{}}
\toprule
\textbf{Top-Level Category} & \textbf{Definition} & \textbf{Representative Secondary Categories} \\
\midrule
Entities & Physical objects, participants, and observation devices with persistent identities that can be tracked within a scene. & Characters, objects, constraint bodies, cameras, lights, and other interactive scene elements \\
Structure & Relatively stable spatial, topological, hierarchical, and assembly relationships within and between entities. & Scene graphs, node hierarchies, skeletons, skins, mesh topology, and assembly relationships \\
Static Properties & Semantic, geometric, material, or physical attributes that remain invariant over their declared validity intervals. & Category, asset ID, mass, friction, material parameters, and other temporally stable attributes \\
Dynamic State & Time-varying entity poses, velocities, joints, animation states, solver states, or internal physical states. & Pose, velocity, joint angles, object states, and internal states of deformable bodies \\
Relations & Interactions between entities that are established, maintained, or terminated over time. & Time-varying contact, support, grasp, visibility, and occlusion relations \\
Events \& Rules & Discrete state transitions and their triggering conditions, together with the mechanisms, constraints, and runtime configurations that govern state evolution. & Semantic events, collisions, constraint switches, fractures, state transitions, and logical triggers \\
Observations & Multimodal outputs derived from the world state by cameras, sensors, renderers, or other observation functions. & RGB, depth, segmentation, normals, optical flow, multiview images, panoramas, and point clouds \\
\bottomrule
\end{tabular}
}
\caption{\changed{Core World-State Data Categories and Representative Contents}}
\label{tab:world-state-categories}
\end{table*}

\subsection{Logical Schema and Sample Definition}

\changed{To enable reuse across toolchains, tasks, and models, we decompose the data-production process and align it with real-world elements.}

Under this formulation, world-state data comprise a layered temporal record of a single world evolution process. Let the canonical time domain of a world segment be $T = \{t_0,t_1,\ldots,t_n\}$. At any time $t$, the explicit world state is

\begin{equation}
W_t = (E, H, P, X_t, R_t, Q_t, \Pi_t).
\end{equation}

Here, $E$ is the entity set; $H$ the structure set; $P$ the static properties; $X_t$ the dynamic state; $R_t$ the time-varying relations; $Q_t$ the events; and $\Pi_t$ the rules and configurations. Given an observation function $g_m$ for modality $m$ and an observation configuration $C_t^m$, the observation at time $t$ is

\begin{equation}
O_t^m = g_m\!\left(W_t; C_t^m, \epsilon_t^m\right).
\end{equation}

The modality configuration $C_t^m$ represents sampling mechanisms such as cameras, scanners, or text interfaces, while $\epsilon_t^m$ represents observation noise or encoding perturbations. Information from each modality is therefore treated as an observation derived from the same world state rather than as the world itself. With a time-varying action $a_t$ and an external intervention $u_t$, world-state evolution is

\begin{equation}
W_{t+1} = F\!\left(W_t, a_t, u_t; \Theta_t\right) + \xi_t.
\end{equation}

Here, $F$ is the state-transition process jointly induced by the rendering engine, physical simulation, camera system, and related components; $\Theta_t$ denotes the solver, rules, and runtime configuration; and $\xi_t$ captures numerical error, stochastic perturbations, or unmodeled factors. When $u_t = \varnothing$, the segment records the original or main trajectory; when $u_t \neq \varnothing$, it records an intervention branch derived from the same or a shared initial state.

A complete CG-World data unit is organized as

\begin{equation}
D_i = \left(W_{t_0:t_n}, O_{t_0:t_n}^{1:M}, A_{t_0:t_n}, U_{t_0:t_n}, B_i, \mathcal{M}_i\right).
\end{equation}

In this tuple, $W_{t_0:t_n}$ is the world-state sequence, $O_{t_0:t_n}^{1:M}$ the multimodal observation sequence, $A_{t_0:t_n}$ the action or control signals, $U_{t_0:t_n}$ the intervention variables, $B_i$ the branch lineage, and $\mathcal{M}_i$ the metadata specification, including fixed information such as coordinate systems, units, and rendering configurations. World model training can therefore extend beyond fitting observations or operational outcomes to jointly learning the next world state:

\begin{equation}
p\!\left(W_{t+1}, O_{t+1}\mid W_{\leq t}, O_{\leq t}, A_{\leq t}, U_{\leq t}\right).
\end{equation}

\begin{figure*}[!t]
\centering
\changedfigure{\includegraphics[width=0.96\textwidth]{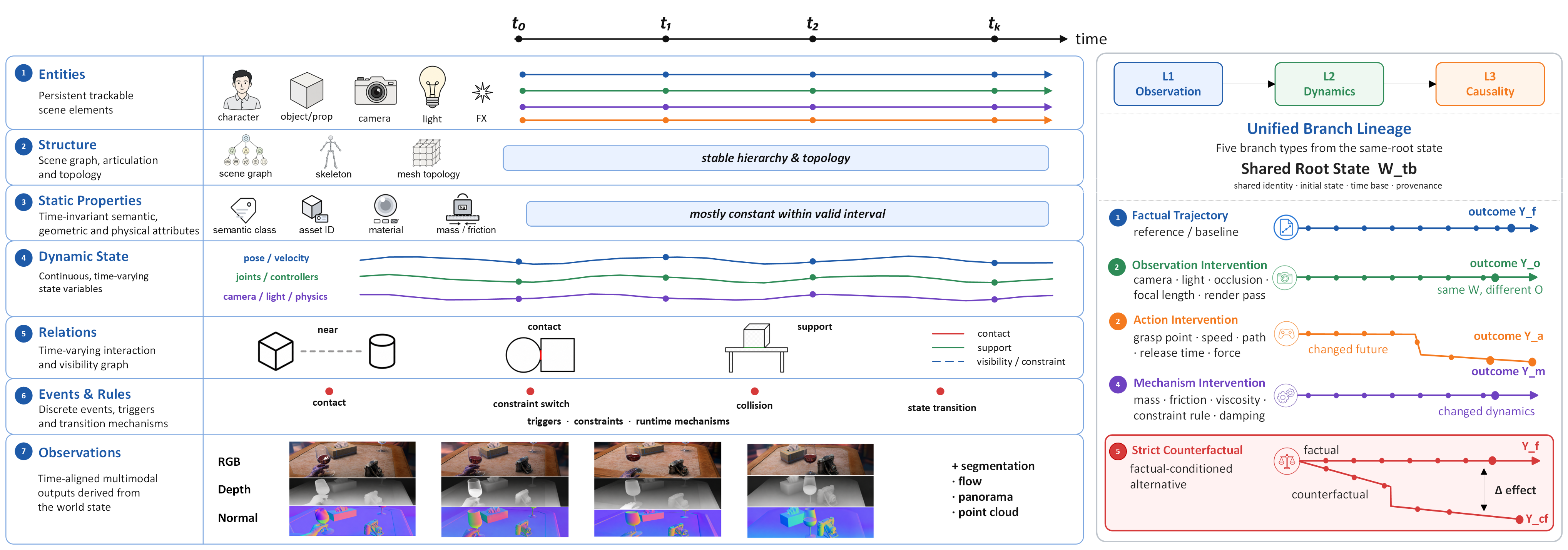}}
\caption{\changed{Time-aligned world-state protocol.}}
\label{fig:protocol}
\end{figure*}

To make the protocol executable, CG-World declares the semantic and data types, tensor shape, temporal index, units, coordinate system, availability, and provenance of every core field. Payloads are referenced through standard files or protocols such as OpenUSD \cite{pixar2025openusd}, RLDS \cite{ramos2021rlds}, and MaterialX; the complete schema is provided in the supplementary material. Each segment is indexed by segment\_id, root\_family\_id, and entity ID and must include a temporal range, entity hierarchy, sensor categories and parameters, and at least one visual observation. Geometry, animation, physics caches, contacts, depth, normals, and optical flow are optional according to source recoverability. As illustrated in Figure~\ref{fig:protocol}, the resulting logical sample layer links world state, actions, state transitions, multimodal observations, relations and events, and causal branches under a unified spatiotemporal index while supporting traceable implementations across industrial tools, simulation platforms, and real-world acquisition systems.

\subsection{Organization Principles and Capability Levels}

We adopt three principles for organizing world-state data:

\begin{enumerate}
\setlength{\itemsep}{0pt}
\setlength{\parsep}{0pt}
\setlength{\topsep}{2pt}
\item Separate states, observations, and annotations, preventing a model from treating compressed observations or human labels as the world itself.
\item Separate templates, instances, and states, allowing changes to the same instance to be tracked over time.
\item Separate continuous states from discrete events while aligning all data to a unified timecode for cross-modal synchronization.
\end{enumerate}

We define three capability levels for world model datasets. L1, the observation-alignment level, minimally requires entity indices, spatial structure or reconstruction references, camera or sensor parameters, basic observations, unified timecodes, coordinate systems, and units. L2, the dynamic-state level, adds action trajectories, entity dynamics, relations, and events. L3, the causal-intervention level, further adds intervention variables, invariant fields, branch lineages, outcome variables, and comparable factual-intervention or factual-counterfactual pairs.

\subsection{Counterfactual and Branch-Management Protocol}

A branch is a world-state trajectory generated from a traceable world state and its runtime context under specific observation conditions, action inputs, or dynamical mechanisms. Following the distinction among factual, interventional, and counterfactual reasoning \cite{pearl2009causality}, CG-World divides branches into factual trajectory, observational intervention, action intervention, mechanism intervention, and strict counterfactual branches.

\noindent\textbf{Factual trajectory branch.} Records the world-state evolution that occurred and was fully preserved under a given runtime context, action input, and mechanism configuration.

\noindent\textbf{Observational intervention branch.} Modifies the observation function from the same initial state while keeping the underlying state trajectory unchanged.

\noindent\textbf{Action intervention branch.} Changes the action or control input applied to the world under a fixed initial world state and runtime context, then records the resulting future states and observations.

\noindent\textbf{Mechanism intervention branch.} Modifies one or a small number of mechanism variables $I$ under the same initial state to obtain a new world state and subsequent trajectory.

\noindent\textbf{Strict counterfactual branch.} Conditions on a factual trajectory that occurred and was fully recorded, then applies an alternative intervention to an action, state, or mechanism variable while holding fixed the corresponding exogenous context and nonintervened mechanisms. CG-World labels a branch as strict counterfactual data only when it satisfies all four criteria: factual conditioning, preservation of the exogenous context, execution of an alternative intervention, and recording of individual-level effects.

\section{CG-World v1}

\subsection{Data Sources and Processing}

\begin{figure*}[!t]
\centering
\changedfigure{\includegraphics[width=0.78\textwidth]{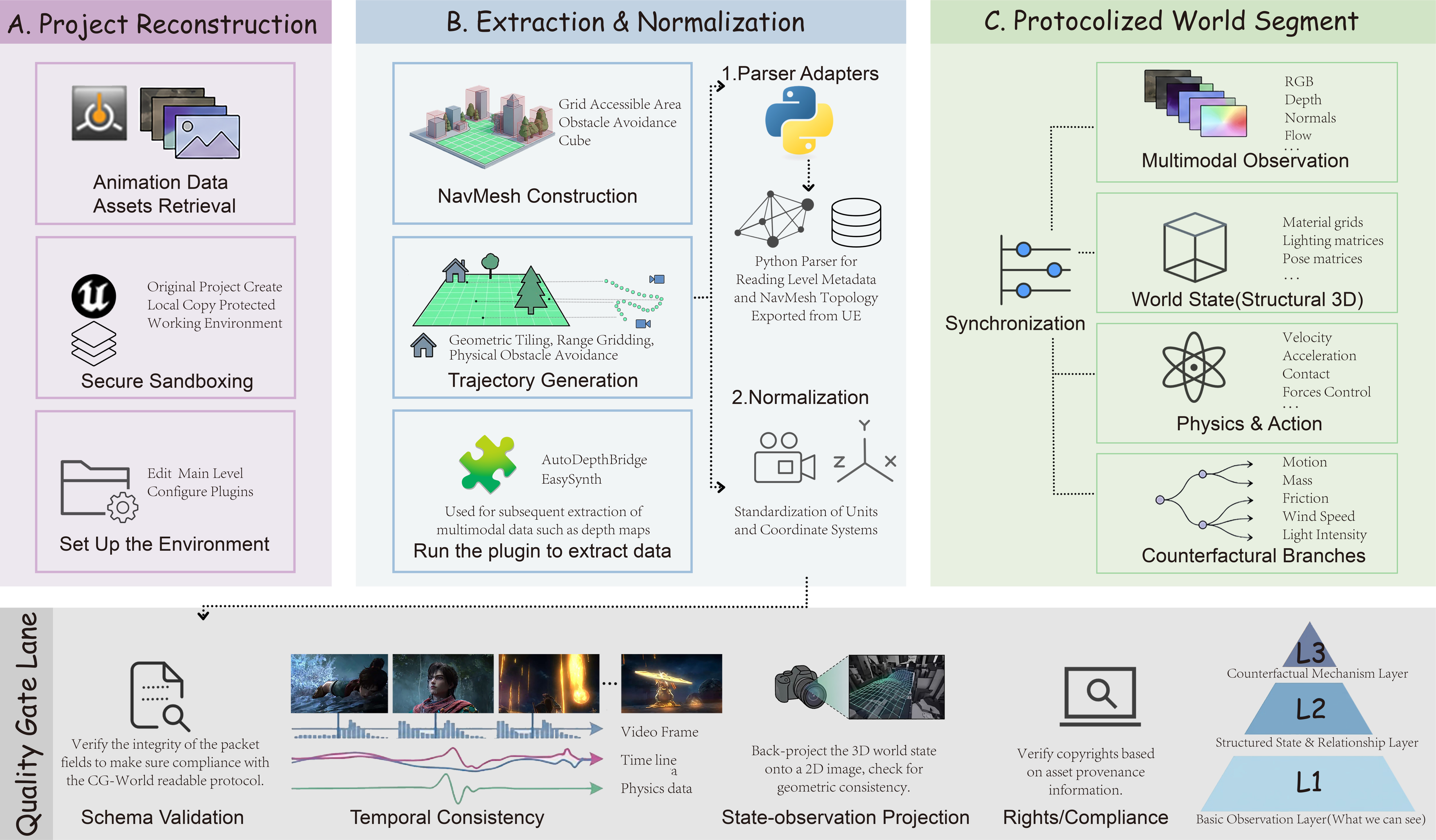}}
\caption{\changed{Processing pipeline from industrial CG projects to CG-World data packages.}}
\label{fig:pipeline}
\end{figure*}

CG-World v1 is a shot-level world-state dataset recovered and standardized from industrial CG production projects authorized for research use and redistribution. We recover temporal correspondences between world states and multiversion rendered observations, using world-state segments with temporal continuity, persistent entity identities, and strict state-observation alignment as the basic sample unit. The pipeline comprises industrial project parsing, cross-toolchain extraction, and protocol-based reorganization, as shown in Figure~\ref{fig:pipeline}.

\subsection{Data Content, Statistical Units, and Scale}

During construction of the CG-World v1 main dataset, we did not augment the data through procedural scene randomization or resimulation. All states and observations in v1 originate from saved production records that passed production quality control.

\begin{figure*}[!t]
\centering
\changedfigure{\includegraphics[width=0.76\textwidth]{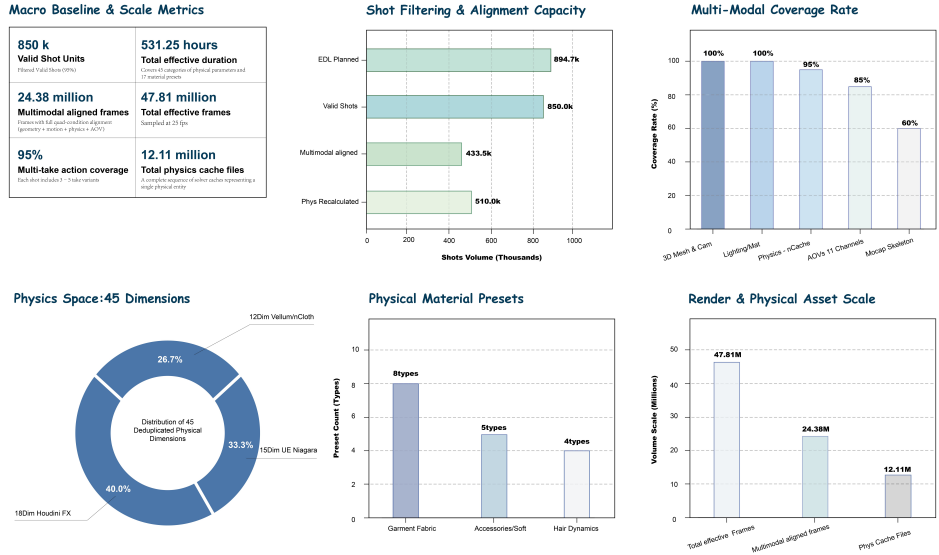}}
\caption{\changed{Data distribution of CG-World v1.}}
\label{fig:distribution}
\end{figure*}

Figure~\ref{fig:distribution} summarizes the full scale, shot-filtering process, and multimodal coverage of CG-World v1. The dataset contains approximately 850,000 valid shot-level world-state segments, each 1-5 seconds long, with a median duration of 2.6 seconds. Of these, 433,500 shots have complete multimodal alignment and 510,000 have physics recalculation. Ninety-five percent of valid shots include multi-take action coverage, with 3-5 take variants per shot. To prevent leakage across production projects, shared assets, historical versions, and same-root branches, CG-World v1 uses grouped hierarchical partitioning. The final training, validation, and test sets contain 637,500, 127,500, and 85,000 segments, respectively, corresponding to 75\%, 15\%, and 10\% of the valid data.

\subsection{Multibranch and Counterfactual Data}

\begin{figure*}[!t]
\centering
\changedfigure{\includegraphics[width=0.90\textwidth]{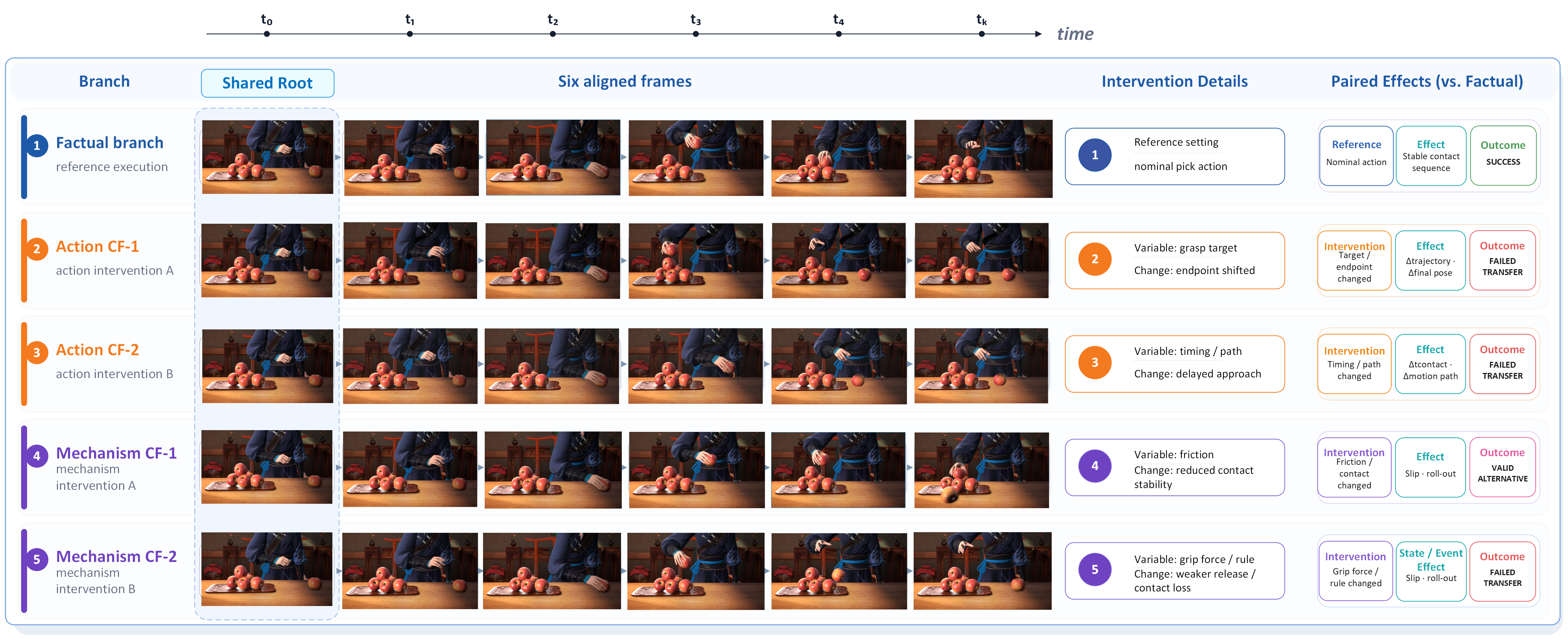}}
\caption{\changed{Counterfactual data structure in CG-World.}}
\label{fig:counterfactual}
\end{figure*}

Using the branch system introduced in Section 3.3, CG-World v1 constructs an intervention and counterfactual subset from desktop container-manipulation scenes. The subset is organized by same-root branch family\changed{, as illustrated in Figure~\ref{fig:counterfactual}}. Each family uses one fully recorded factual trajectory as its reference and associates it with alternative branches derived from the same traceable world state.

Approximately 9\% of this subset is derived from historical project versions, with the remainder generated procedurally. Table~\ref{tab:branches} summarizes its principal organizational units, branch scales, intervention targets, and derivable supervision.

\begin{table*}[!t]
\centering
\footnotesize
\setlength{\tabcolsep}{2.5pt}
\renewcommand{\arraystretch}{0.98}
\begin{tabular}{@{}L{0.19\textwidth}L{0.10\textwidth}L{0.27\textwidth}L{0.29\textwidth}@{}}
\toprule
\textbf{Data Unit or Branch Type} & \textbf{Scale} & \textbf{Primary Variables Changed} & \textbf{Supervision Provided} \\
\midrule
Same-root branch family & 1,000 groups & Shared factual trajectory, initial state, runtime context, and branch lineage & Factual reference, branch correspondences, and invariant sets \\
Observational intervention branch & 100,000 & Camera, focal length, lighting, occlusion, render passes, and image degradation & Multi-pass observations, visibility, camera parameters, and observation differences \\
Strict action counterfactual branch & 5,000 & Grasp position, speed, tilt angle, motion path, and release timing & Action trajectories, future states, event boundaries, and individual-level outcome effects \\
Strict mechanism counterfactual branch & 5,000 & Mass, friction, center of mass, liquid level, viscosity, contact conditions, and constraint rules & Physical states, contact relations, event changes, and individual-level outcome effects \\
Failure-outcome annotations & Included in the corresponding counterfactual branches & Slippage, spillage, collision, premature release, and related failures & Failure events, anomalous contacts, occurrence times, and final-state changes \\
\bottomrule
\end{tabular}
\caption{\changed{Branch and Counterfactual Subset of CG-World v1}}
\label{tab:branches}
\end{table*}

\subsection{Annotation and Organization}

CG-World v1 uses production-derived annotation: annotations are recovered from industrial production files rather than manually redrawn as boxes, masks, or textual descriptions over extracted segments. Textual semantics, structure, actions, physical processes, and observations are produced within the same production chain under human expert supervision.

\subsection{Data Rights, Licensing, and Release Plan}

The official CG-World v1 release includes only industrial CG projects that have passed rights review and permit academic research use. Materials that do not satisfy redistribution requirements are used only for internal statistics or controlled access and are excluded from the public download.

The complete CG-World v1 dataset will be released upon paper acceptance and no later than camera-ready submission. L1/L2 and selected L3 data will be distributed under a noncommercial academic research license. Raw L3 data, comprising the complete project files corresponding to world-state segments, will require a use-case review and a signed controlled-access agreement.

\section{Experiments}

We conduct three experiments: observation video generation, multi-step action prediction, and vision-language-action transfer. \changed{Table~\ref{tab:models} summarizes the evaluated models and initialization checkpoints.}

\begin{table}[ht]
\centering
\footnotesize
\setlength{\tabcolsep}{2.5pt}
\renewcommand{\arraystretch}{1.00}
\begin{tabular}{@{}L{0.28\columnwidth}L{0.31\columnwidth}L{0.25\columnwidth}@{}}
\toprule
\textbf{Task} & \textbf{Model and Version} & \textbf{Post-Training Method} \\
\midrule
Observation video generation & ltx-2.3-22b-dev \cite{hacohen2026ltx2} & LoRA \cite{hu2022lora}, r = 32 \\
Multi-step action prediction & Cosmos3-Nano \cite{nvidia2026cosmos3} & LoRA, r = 32 \\
Embodied policy transfer & openvla-7b & LoRA, r = 32 \\
\bottomrule
\end{tabular}
\caption{\changed{Experimental Models and Initialization Checkpoints}}
\label{tab:models}
\end{table}

\subsection{Observation Video Generation}

We first evaluate whether supervision derived from world state improves observation generation in a video foundation model. Four settings are compared under the same initial frame, prompt, test clip, random seed, and inference configuration: I2V without explicit control, Union Control LoRA \cite{benyosef2026avcontrol}, CG-World single-modality post-training, and mixed Depth/Canny post-training.

\begin{center}
\begin{minipage}{\columnwidth}
\centering
\footnotesize
\setlength{\tabcolsep}{2.5pt}
\renewcommand{\arraystretch}{1.00}
\begin{tabular}{@{}L{0.14\columnwidth}L{0.43\columnwidth}R{0.15\columnwidth}R{0.15\columnwidth}@{}}
\toprule
\textbf{Control} & \textbf{Method} & \textbf{LPIPS $\downarrow$} & \textbf{FVD $\downarrow$} \\
\midrule
Depth & LTX-2.3 baseline I2V & 0.503639 & 2568.610 \\
Depth & + Official Union & 0.446168 & 532.730 \\
Depth & + CG-World Depth & 0.406837 & 519.624 \\
Depth & + Mixed Union & 0.278523 & 326.324 \\
Canny & LTX-2.3 baseline I2V & 0.503639 & 2568.610 \\
Canny & + Official Union & 0.454424 & 652.224 \\
Canny & + CG-World Canny & 0.398862 & 555.545 \\
Canny & + Mixed Union & 0.304219 & 368.162 \\
\bottomrule
\end{tabular}
\captionof{table}{\changed{Post-Training Results of LTX-2.3 on Observation Video Generation}}
\label{tab:video-results}
\end{minipage}
\end{center}

We use LPIPS \cite{zhang2018lpips} and FVD \cite{unterthiner2018fvd} to measure perceptual and distributional distances between generated and real videos. As shown in Table~\ref{tab:video-results}, CG-World geometry-control supervision yields the best results in both control branches. Relative to Official Union, LPIPS and FVD decrease by 37.57\% and 38.75\% in the Depth branch and by 33.05\% and 43.55\% in the Canny branch. These results indicate that strict alignment among same-source RGB, geometric controls, and temporal information improves perceptual fidelity and distributional consistency in controlled video generation.

\subsection{Multi-Step Action Prediction for Physical AI}

The second experiment evaluates whether CG-World trajectory supervision improves world model action prediction. The data are organized according to LIBERO \cite{liu2023libero}, and poses use a continuous six-dimensional rotation representation \cite{zhou2019rotation}. Using Cosmos3-Nano as the baseline, we construct three controls with a shared action connector and prediction head. Action-Head Only freezes the model backbone and optimizes only the action head and connector; LIBERO-LoRA post-trains the backbone on LIBERO; and CG-World-LoRA post-trains it on CG-World. Given the current RGB observation and a natural-language instruction, the model predicts a 16-step future action chunk. All three configurations share the same action head, data-partitioning principles, training budget, and frozen evaluation protocol.

\begin{center}
\begin{minipage}{\columnwidth}
\centering
\footnotesize
\setlength{\tabcolsep}{2.5pt}
\renewcommand{\arraystretch}{1.00}
\begin{tabular}{@{}L{0.30\columnwidth}C{0.18\columnwidth}C{0.18\columnwidth}C{0.18\columnwidth}@{}}
\toprule
\textbf{Metric} & \textbf{Action-Head Only} & \textbf{LIBERO-LoRA} & \textbf{CG-World-LoRA} \\
\midrule
Action MSE $\downarrow$ & 0.11532 & 0.08061 & 0.05326 \\
Translation RMSE $\downarrow$ & 0.41624 & 0.34084 & 0.27152 \\
Rotation Error ($^\circ$) $\downarrow$ & 3.88405 & 4.53336 & 2.13520 \\
Gripper F1 $\uparrow$ & 0.70314 & 0.89354 & 0.95726 \\
\bottomrule
\end{tabular}
\captionof{table}{\changed{Cosmos3-Nano Action-Prediction Results}}
\label{tab:action-results}
\end{minipage}
\end{center}

As shown in Table~\ref{tab:action-results}, CG-World-LoRA achieves the best result on all four action metrics. Relative to LIBERO-LoRA, the three continuous-action errors decrease by 33.93\%, 20.34\%, and 52.90\%, while Gripper F1 improves by 0.06372 in absolute terms. These results support the claim that CG-World provides continuous motion structure beyond independent demonstration trajectories.

\subsection{VLA Transfer in Simulation}

The third experiment evaluates whether CG-World provides transferable motion priors for a VLA model and whether those priors improve closed-loop manipulation under limited target-domain supervision. We use OpenVLA-7B with the OFT continuous-action regression paradigm \cite{kim2025oft}. Given an RGB observation and task instruction, the model outputs a 7D relative end-effector action. We compare post-training with ManiSkill3 \cite{tao2024maniskill3} and CG-World. The CG-World pre-adapt + OFT condition adds CG-World action pre-adaptation before the same target-domain OFT protocol. We evaluate in-domain task success on PickCube, cross-task generalization on the unseen StackCube task, and decompose failures into Grasp and Lift stages.

\changed{As shown in Table~\ref{tab:vla-results},} CG-World pretraining raises the in-domain PickCube success rate from 4.0\% for C2 to 39.0\% and improves StackCube OOD success from 0\% for every baseline to 25.0\%. Its primary benefit is not further reinforcement of already saturated target approach and grasping, but improved post-grasp lifting, trajectory continuity, object transport, and subsequent action planning.

\begin{center}
\begin{minipage}{\columnwidth}
\centering
\scriptsize
\setlength{\tabcolsep}{2pt}
\renewcommand{\arraystretch}{1.08}
\begin{tabular}{@{}L{0.26\columnwidth}C{0.17\columnwidth}C{0.22\columnwidth}C{0.17\columnwidth}@{}}
\toprule
\textbf{Configuration} & \textbf{PickCube ID SR} & \textbf{Wilson 95\% CI} & \textbf{StackCube OOD SR} \\
\midrule
C1 Zero-shot & 0.7 $\pm$ 1.2\% & 0.7\% [0.2\%, 2.4\%] & 0.0 $\pm$ 0.0\% \\
C2 RGB BC (OFT) & 4.0 $\pm$ 4.4\% & 4.0\% [2.3\%, 6.9\%] & 0.0 $\pm$ 0.0\% \\
CG-World pre-adapt + OFT & 39.0 $\pm$ 4.4\% & 39.0\% [33.7\%, 44.6\%] & 25.0 $\pm$ 4.4\% \\
\bottomrule
\end{tabular}
\vspace{3pt}
\begin{tabular}{@{}L{0.44\columnwidth}C{0.21\columnwidth}C{0.21\columnwidth}@{}}
\toprule
\textbf{Configuration} & \textbf{Grasp} & \textbf{Lift} \\
\midrule
C1 Zero-shot & 0.0 $\pm$ 0.0\% & 0.0 $\pm$ 0.0\% \\
C2 RGB BC (OFT) & 99.0 $\pm$ 1.0\% & 68.7 $\pm$ 7.5\% \\
CG-World pre-adapt + OFT & 99.5 $\pm$ 0.5\% & 88.3 $\pm$ 3.1\% \\
\bottomrule
\end{tabular}
\captionof{table}{\changed{OpenVLA Simulation Results on Desktop Container-Manipulation Transfer}}
\label{tab:vla-results}
\end{minipage}
\end{center}

\section{Future Work}

\changed{Future work will use strictly registered real--virtual scene pairs to measure biases in observations, dynamics, and intervention effects; develop error and uncertainty models; and calibrate industrial CG mechanisms and sampling ranges with limited real-world data. We will also expand long-horizon causal chains and dynamical coverage. Community plugins will map projects to the CG-World schema, while contributions will require schema, temporal-consistency, authorization, and provenance checks and receive version and citation metadata under the L1/L2/L3 hierarchy.}

\section{Discussion and Conclusion}

\subsection{Discussion}

\changed{CG-World v1 inherits production biases: narrative and shot selection, stylized or corrected motion, engine- and solver-specific behavior, incomplete legacy records, and weakened state--observation correspondence from compositing, transparency, or procedural materials. Its production-derived states should therefore be used as explicit supervision with documented provenance and operating conditions, not as uncalibrated real-world physical ground truth.}

\subsection{Conclusion}

\changed{CG-World aligns industrial-CG states, actions, multimodal observations, and causal branches through a world-state protocol and large-scale dataset. This supervision supports modeling state evolution, observation generation, and intervention outcomes beyond terminal-pixel fitting, improving controlled video generation, action prediction, and embodied manipulation transfer.}

\bibliography{cgworld}

\end{document}